\documentclass[sigconf]{acmart}

\usepackage{booktabs}
\usepackage{multirow}
\usepackage{colortbl}
\usepackage{algorithm}
\usepackage{algorithmic}
\usepackage{tikz}
\usetikzlibrary{arrows.meta,positioning,shapes.geometric,calc}

\setcopyright{rightsretained}
\copyrightyear{2026}
\acmYear{2026}
\acmDOI{}
\acmConference[AI4Finance Summit @ ICAIF '26]{AI4Finance Summit at ICAIF '26 (7th ACM International Conference on AI in Finance)}{November 14--17, 2026}{Milan}
\acmBooktitle{AI4Finance Summit at ICAIF '26 (7th ACM International Conference on AI in Finance), November 14--17, 2026, Milan}
\acmISBN{}

\begin{document}

\title{Could This BE Any More Wrong? Teaching VLMs to Know What They Don't Know in Document Extraction}

\author{Varad Srivastava\textsuperscript{*}, Victor Casas Hernandez\textsuperscript{\dag}}
\affiliation{\institution{{\Large AI Lab, Barclays Investment Bank}}%
\institution{\vspace{4pt}\texttt{\{\textsuperscript{*}varad.srivastava, \textsuperscript{\dag}victor.casashernandez\}@barclays.com}}%
\country{}}
\authorsaddresses{}

\renewcommand{\shortauthors}{Srivastava and Casas Hernandez}

\begin{abstract}
Vision-Language Models can extract structured fields from scanned financial
documents, yet they provide no indication of when a field might be wrong.
We derive per-field confidence estimates from the token log-probabilities
the model already produces, via a token-to-span alignment procedure that maps
logprobs onto each extracted value within the structured JSON output.
Evaluated on two benchmarks (CORD~v2 receipts, Kleister~NDA legal contracts)
with three VLM families, the minimum token probability per field achieves
Area Under the ROC Curve (AUROC) 0.74--0.91 for error detection---far exceeding verbalized confidence
(0.53--0.56).
The uncertain fields form a natural training curriculum: fine-tuning via Group
Relative Policy Optimisation (GRPO) improves accuracy from 78.5\% to 87.5\% on
CORD (67 examples) and from 70.7\% to 75.1\% on multi-page NDA contracts (259
examples), while substantially improving error detection on Kleister NDA
(AUROC~0.70$\to$0.85, Area Under the Accuracy--Rejection Curve (AUARC)~0.78$\to$0.92).
The loop requires no annotation beyond the training-split ground truth,
demonstrating that confidence-guided curriculum selection transfers to complex
multi-page documents when combined with native multi-image training.
A frontier model (Claude Sonnet~4) used as a pseudo-label judge achieves
${\sim}$92\% reward agreement with ground truth on both benchmarks,
suggesting this loop can operate without human annotation.
\end{abstract}



\maketitle

\section{Introduction}

Straight-through processing of financial documents (trade confirmations,
invoices, regulatory filings) is a long-standing goal of operations technology
in banking and insurance.
Recent Vision-Language Models (VLMs) have made it feasible: given a scanned
image, a single inference call can produce a structured JSON object containing
all relevant fields (amounts, dates, counterparties, reference numbers).
On well-formatted documents, extraction accuracy routinely exceeds 80\%.

The practical problem is not accuracy in the aggregate but reliability at the
field level.
A trade confirmation where the notional amount is wrong but the settlement date
is correct still requires human intervention, but only on the notional field.
Routing entire documents to manual review when a single field is uncertain wastes
the efficiency gains that motivated VLM adoption.
What is needed is a per-field confidence signal: a score attached to each
extracted value that predicts whether that specific extraction is likely correct.

Existing approaches to this problem have significant drawbacks in the financial
document setting.
\emph{Verbalized confidence}~\cite{yang2024verbalized}, where the model is asked
to rate its own certainty, is known to be poorly calibrated and prompt-dependent;
we confirm empirically that it achieves AUROC of only 0.53--0.56 (near random) on our
benchmarks.
\emph{Fine-tuned calibrators}~\cite{xiao2026vlm} require labelled training data
from the target domain, which is expensive to produce for proprietary financial
documents.
\emph{Sampling-based methods}~\cite{manakul2023selfcheckgpt, kuhn2023semantic}
require multiple forward passes per document, which is incompatible with
throughput requirements in production document processing pipelines.

Our starting observation is simple: VLMs already produce token-level
log-probabilities during generation at no additional cost.
These probabilities reflect the model's uncertainty about each generated
token, but the information that matters for downstream routing is at the
\emph{field} level.
That token logprobs are informative is not itself new~\cite{yuan2024fact,
tao2025uncertainty}; our contribution is the complete system that makes them
\emph{actionable at field granularity} for structured document extraction.
Concretely, we contribute:
(i)~a token-to-span alignment procedure that handles the non-trivial
character mapping in JSON with escaped values, nested arrays, and byte-pair
encoding (BPE) tokenisation;
(ii)~an empirical demonstration across three model families
(Qwen3-VL-8B, Granite-2.9B, Gemma~4-5.1B) that
\texttt{min\_prob} is consistently discriminative for this task; and
(iii)~a confidence-guided self-improvement loop via GRPO that improves
both accuracy and calibration on single-page and multi-page documents,
together with an analysis of the conditions under which this loop succeeds
and evidence that frontier-model pseudo-labels can replace ground-truth
rewards.

In our experiments, \texttt{min\_prob} achieves AUROC 0.74--0.91 across all
three model families and both domains, with no training required.
The uncertain fields form a natural curriculum: GRPO fine-tuning lifts
accuracy by +9pp on CORD (67 examples) and +4.4pp on multi-page NDA
contracts (259 examples, AUROC 0.70$\to$0.85), though curriculum volume
and document diversity prove critical on NDA
(Section~\ref{sec:nda-grpo}).
A frontier model used as pseudo-label judge achieves ${\sim}$92\% reward
agreement with ground truth on both benchmarks, suggesting this loop can
operate without human annotation (Section~\ref{sec:pseudo}).

\section{Related Work}

\paragraph{Token probabilities as confidence.}
The idea of using generation probabilities for uncertainty estimation is not new.
\citet{kadavath2022lm} show that language models ``mostly know what they know''
when probed via P(True), though their method requires sampling multiple candidate
answers.
\citet{tao2025uncertainty} conduct a systematic comparison of token-probability
uncertainty (TPU), numerical verbal uncertainty, and linguistic verbal uncertainty
across 80 LLMs, finding that TPU provides competitive discrimination on reasoning
tasks.
Most relevant to our work, \citet{snel2025firsttoken} find that the first
hallucinated token in a generated span is significantly more detectable (via its
logprob) than subsequent tokens, a structural property that explains why
\texttt{min\_prob} outperforms the arithmetic mean in our experiments.

\citet{chuang2024confidence} train dedicated confidence tokens into LLMs for
routing decisions; while effective, this requires task-specific training data.
Our method requires no task-specific training data.

\paragraph{Calibration of neural networks and LLMs.}
\citet{guo2017calibration} establish that modern neural networks are miscalibrated
and introduce temperature scaling as a post-hoc fix.
For LLMs specifically, \citet{nakkiran2025rl} find that RLHF-style instruction
tuning disrupts semantic calibration, yet \citet{yang2025concentration} show
that alignment \emph{concentrates} probability mass, making the resulting logprobs
more informative for uncertainty detection despite poorer raw calibration.
\citet{chen2026greedy} provide a partial answer: greedy decoding produces
better-calibrated probabilities than sampling for VQA tasks, motivating our
design choice to enforce greedy decoding in the confidence pipeline.

\paragraph{Confidence for document IE.}
\citet{yuan2024fact} propose fact-level confidence by aggregating token logprobs
over named-entity spans in free-text generation.
Our work extends this idea to the VLM setting where the output is structured JSON
containing quoted, escaped values with non-trivial character-span recovery.
\citet{nguyen2022hycedis} approach the same problem differently, training a
separate conformal predictor and variational anomaly detector on top of a document
extraction model; our method requires no additional learned components.

\paragraph{VLM-based document extraction.}
Layout-aware models such as LayoutLM~\cite{xu2020layoutlm} established strong
baselines by jointly pre-training text and spatial embeddings.
\citet{kim2022donut} demonstrated OCR-free document parsing with an
encoder-decoder transformer.
More recently, general-purpose VLMs match fine-tuned models via prompting
alone~\cite{gomez2026invoices}, achieving F1\,{>}\,96\% on invoices without
task-specific training.
These results establish VLMs as competitive extractors; the open
problem, which we address here, is knowing \emph{when} to trust the output.

\paragraph{Self-improvement in LLMs.}
STaR~\cite{zelikman2022star} bootstraps reasoning by fine-tuning on
rationales that led to correct answers.
ReST~\cite{gulcehre2023rest} formalises a generate-filter-train loop.
DeepSeek-R1~\cite{deepseek2025r1} shows RL on verifiable rewards elicits
strong reasoning without human trajectories.
\citet{ding2024gsi} identify \emph{tail narrowing}: easy examples dominate
and hard cases starve.
Our uncertainty-based selector addresses this directly: fields with the
lowest confidence scores are selected first, though the optimal signal
for selection depends on field length (Section~\ref{sec:curriculum}).

\section{Method}

\subsection{Overview}

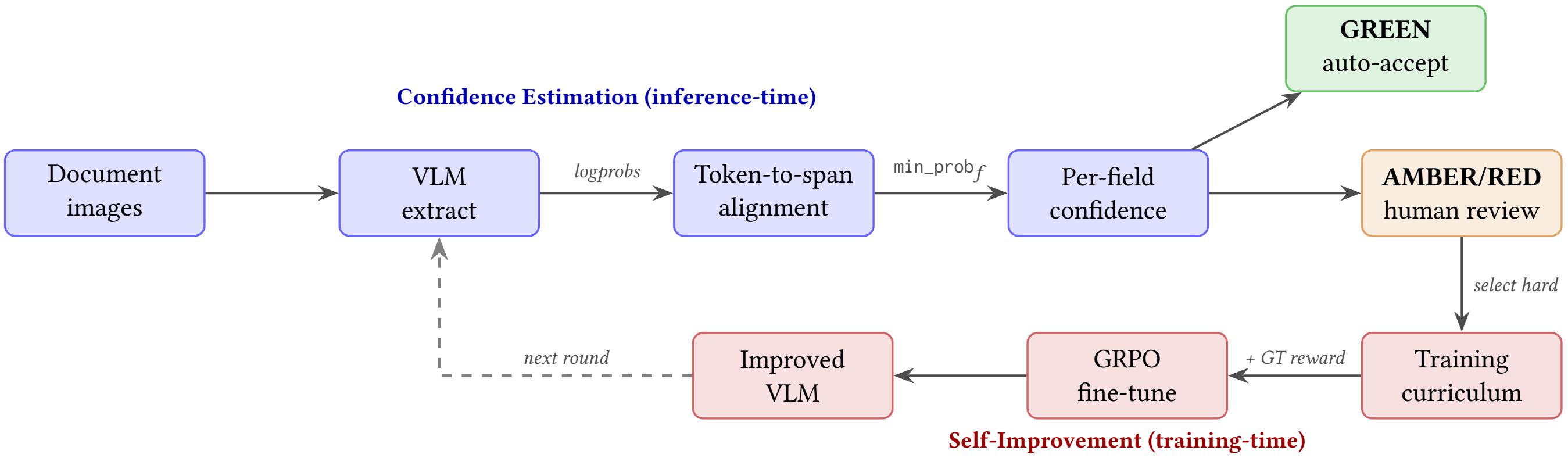
\begin{figure*}[t]
\centering
\resizebox{\textwidth}{!}{%
\begin{tikzpicture}[
    >=Stealth,
    node distance=1.1cm and 1.4cm,
    box/.style={rectangle, draw, rounded corners=3pt, minimum height=0.9cm,
                minimum width=2.1cm, align=center, font=\small,
                fill=#1!12, draw=#1!60, line width=0.6pt},
    box/.default=black,
    lbl/.style={font=\scriptsize\itshape, text=black!70},
    arr/.style={->, line width=0.7pt, color=black!70},
    darr/.style={->, line width=0.9pt, dashed, color=black!50},
  ]
  \node[box=blue] (doc) {Document\\images};
  \node[box=blue, right=of doc] (vlm) {VLM\\extract};
  \node[box=blue, right=of vlm] (logp) {Token-to-span\\alignment};
  \node[box=blue, right=of logp] (conf) {Per-field\\confidence};
  \node[box=green!60!black, above right=0.6cm and 0.8cm of conf] (green) {\textbf{GREEN}\\auto-accept};
  \node[box=orange!80!black, right=1.6cm of conf] (amberred) {\textbf{AMBER/RED}\\human review};

  \node[box=red!70!black, below=1.0cm of amberred] (curriculum) {Training\\curriculum};
  \node[box=red!70!black, left=of curriculum] (grpo) {GRPO\\fine-tune};
  \node[box=red!70!black, left=of grpo] (improved) {Improved\\VLM};

  \draw[arr] (doc) -- (vlm);
  \draw[arr] (vlm) -- node[above, lbl] {logprobs} (logp);
  \draw[arr] (logp) -- node[above, lbl] {$\texttt{min\_prob}_f$} (conf);
  \draw[arr] (conf) -- (green);
  \draw[arr] (conf) -- (amberred);

  \draw[arr] (amberred) -- node[right, lbl] {select hard} (curriculum);
  \draw[arr] (curriculum) -- node[above, lbl] {+ GT reward} (grpo);
  \draw[arr] (grpo) -- (improved);
  \draw[darr] (improved) -- node[above, lbl] {next round} (vlm |- improved) -- (vlm);

  \node[font=\footnotesize\bfseries, text=blue!70!black] at ($(doc)!0.5!(conf) + (0, 1.0)$) {Confidence Estimation (inference-time)};
  \node[font=\footnotesize\bfseries, text=red!60!black] at ($(improved)!0.5!(curriculum) + (0, -0.7)$) {Self-Improvement (training-time)};
\end{tikzpicture}%
}
\caption{Pipeline overview.
\textbf{Top:} A single greedy VLM call produces structured JSON with token
logprobs; the confidence stage aligns logprobs to field spans and routes each
field to GREEN (auto-accept) or AMBER/RED (human review).
\textbf{Bottom:} The uncertain fields form a training curriculum for GRPO
fine-tuning, closing a self-improvement loop.
Dashed arrow indicates an optional next iteration.}
\label{fig:pipeline}
\Description{Pipeline diagram showing document image flowing through VLM extraction, logprob-based confidence scoring, RED/AMBER/GREEN routing, and an optional GRPO self-improvement loop.}
\end{figure*}

Figure~\ref{fig:pipeline} illustrates the full pipeline.
The confidence estimation stage (left) operates on a single greedy inference
call with no model changes; the self-improvement stage (right) uses the
resulting uncertainty signal to select a training curriculum.

Algorithm~\ref{alg:main} gives the full procedure, which operates on the output
of a single greedy VLM inference call.
No changes to the model architecture, weights, or inference procedure are
required; the entire confidence computation is post-hoc.

\begin{algorithm}[t]
\caption{Field-Level Confidence Estimation}
\label{alg:main}
\small
\begin{algorithmic}[1]
\REQUIRE Image $I$, prompt $\mathcal{P}$, fields $\mathcal{F}$, top-$k$
\ENSURE Per-field ratings $\{\text{rag}_f\}_{f \in \mathcal{F}}$
\STATE Attach \texttt{LogprobCapture}$(k)$ hook to VLM
\STATE $R, \mathbf{L} \leftarrow \text{VLM.generate}(I, \mathcal{P}, \texttt{greedy})$
\STATE \textit{// Token-to-character alignment}
\FOR{$t = 1$ to $T$}
  \STATE $\text{span}[t] \leftarrow [|D_{t-1}|,\; |D_t|)$ \quad\textit{// prefix decode}
\ENDFOR
\STATE Strip thinking prefix; let answer $A$ start at offset $\delta$
\STATE $\text{fspan} \leftarrow \textsc{JsonSpanScanner}(A)$ \quad\textit{// $f \mapsto [s_f, e_f)$}
\STATE \textit{// Per-field confidence}
\FOR{each $f \in \mathcal{F}$}
  \STATE $\mathcal{T}_f \leftarrow \{t : \text{span}[t] \cap [s_f{+}\delta,\, e_f{+}\delta) \neq \emptyset,\; \mathbf{L}[t] \neq \text{NaN}\}$
  \IF{$\mathcal{T}_f = \emptyset$}
    \STATE $\text{rag}_f \leftarrow \textsc{Red}$; \textbf{continue}
  \ENDIF
  \STATE $\texttt{geo\_prob}_f \leftarrow \exp\bigl(\frac{1}{|\mathcal{T}_f|}\sum_{t}\log p_t\bigr)$
  \STATE $\texttt{min\_prob}_f \leftarrow \min_{t \in \mathcal{T}_f} p_t$
  \STATE Assign $\text{rag}_f$ via thresholds (Section 3.3)
\ENDFOR
\RETURN $\{\text{rag}_f\}_{f \in \mathcal{F}}$
\end{algorithmic}
\end{algorithm}

\begin{enumerate}
\item \textbf{Logprob capture.} A \texttt{LogitsProcessor} hook records the
      top-$k$ token probabilities at each generation step. We use $k{=}50$;
      under greedy decoding the chosen token is nearly always within this window.
\item \textbf{Token-to-character alignment.} Incremental prefix decoding
      maps each generated token to its character interval
      $[\text{start}_t, \text{end}_t)$ in the decoded string.
\item \textbf{JSON span scanning.} A recursive-descent parser identifies the
      character span $[s_f, e_f)$ of each field value in the JSON output.
\item \textbf{Per-field statistics.} For each field $f$, the overlapping tokens
      $\mathcal{T}_f$ yield confidence statistics (defined below).
\item \textbf{Routing.} Thresholds on \texttt{geo\_prob} and \texttt{min\_prob}
      assign each field a GREEN/AMBER/RED label.
\end{enumerate}

\subsection{Confidence Statistics}

For tokens $\mathcal{T}_f$ overlapping field $f$, with $p_t$ the probability
assigned to the chosen token at step $t$:
\begin{align}
\texttt{geo\_prob}_f &= \exp\!\left(\frac{1}{|\mathcal{T}_f|}\sum_{t \in \mathcal{T}_f} \log p_t\right) \\
\texttt{min\_prob}_f &= \min_{t \in \mathcal{T}_f}\; p_t \\
\texttt{avg\_prob}_f &= \frac{1}{|\mathcal{T}_f|}\sum_{t \in \mathcal{T}_f} p_t \\
\texttt{avg\_entropy}_f &= \frac{1}{|\mathcal{T}_f|}\sum_{t \in \mathcal{T}_f} H_t
\end{align}
where $H_t = -\sum_{v \in \text{top-}k} p_t(v) \log p_t(v)$ is the entropy of the
captured distribution at step $t$.
Tokens that fall outside the top-$k$ window (receiving NaN logprob) are excluded.

\subsection{Routing Thresholds}

We assign each field a traffic-light label based on thresholds
calibrated on a separate financial document corpus:

\smallskip
\noindent
\begin{tabular}{@{}ll@{}}
GREEN: & \texttt{geo\_prob}$\,{\geq}\,$0.99 \emph{and} \texttt{min\_prob}$\,{\geq}\,$0.80 \\
AMBER: & \texttt{geo\_prob}$\,{\geq}\,$0.95 \emph{and} \texttt{min\_prob}$\,{\geq}\,$0.40 \\
RED: & all other cases (incl.\ parse failures)
\end{tabular}
\smallskip

\noindent Both conditions are conjunctive: a field needs high average confidence
\emph{and} no single weak token to qualify as GREEN.
The thresholds are stable under perturbation: varying \texttt{geo\_prob} from 0.97 to
0.999 (with \texttt{min\_prob}=0.80 fixed) shifts GREEN accuracy smoothly from
0.886 to 0.936 while GREEN count drops from 236 to 218, with no cliff effects.
The \texttt{min\_prob} threshold has negligible effect above 0.60 (all fields
passing \texttt{geo\_prob}$\geq$0.99 already have \texttt{min\_prob}$\geq$0.80).
These thresholds transfer across domains without recalibration
(Section~\ref{sec:cross}), though Platt scaling can further improve precision
when domain-specific labels are available (Section~\ref{sec:recal}).

\subsection{Confidence-Guided Self-Improvement}
\label{sec:self}

The RED and AMBER fields identified by the routing stage represent the
model's hardest cases---exactly the examples where targeted fine-tuning
helps most.
Rather than simply routing them to human review, they form a natural training
curriculum: the extraction ground truth from the training split serves as the
supervision signal, and the confidence pipeline selects which examples to
train on, closing a self-improvement loop that requires no
\emph{additional} annotation.

We fine-tune using Group Relative Policy Optimisation (GRPO)~\cite{shao2024deepseekmath}, which generates $k{=}4$ completions
per prompt and updates toward higher-reward completions.
Our reward function assigns $+1.0$ for exact match, $+0.5$ for partial
token F1 (character-overlap F1 between predicted and gold strings)~${\geq}$~0.7, and $-0.3$ for invalid JSON.
All fine-tuning uses rank-16 quantized low-rank adaptation (QLoRA) for 3~epochs with $\beta{=}0$ (no KL
penalty; the LoRA constraint itself regularises).
We also compare against standard supervised fine-tuning (SFT) on CORD.

Unlike STaR~\cite{zelikman2022star} and ReST~\cite{gulcehre2023rest},
which filter by correctness, our curriculum selector is the model's own
uncertainty---available on any document without labels.
The key operational insight is that in production deployments, the human
review queue already exists for quality assurance; the corrected fields it
produces \emph{are} the next batch of training data, requiring no separate
annotation effort.

Before computing confidence statistics, the raw model output must be
cleaned of reasoning tokens that some models prepend.

\subsection{Handling Thinking-Mode Models}
\label{sec:thinking}

Several modern VLMs ship with chain-of-thought enabled by default (including
the Thinking variant of Qwen3-VL-8B used in our experiments); we suppress it
via chat-template patching to preserve logprob fidelity.
Residual thinking tokens that occasionally leak through are enclosed in
\texttt{<think>...</think>} markers.
The confidence pipeline handles this by searching for the closing
\texttt{</think>} tag, recording its character offset $\delta$, and computing
span alignment only over post-$\delta$ answer tokens.
If the tag is absent (non-thinking models), $\delta{=}0$ and the full output is
used.
If the tag is malformed or partial (e.g., generation truncated mid-thought), the
entire output is treated as answer text. This conservative fallback may
slightly degrade confidence quality but never produces an error.

\begin{figure*}[t]
\centering
\begin{minipage}[t]{0.48\textwidth}
\centering
\setlength{\fboxsep}{0pt}\setlength{\fboxrule}{0.4pt}%
\fbox{\includegraphics[height=5.8cm]{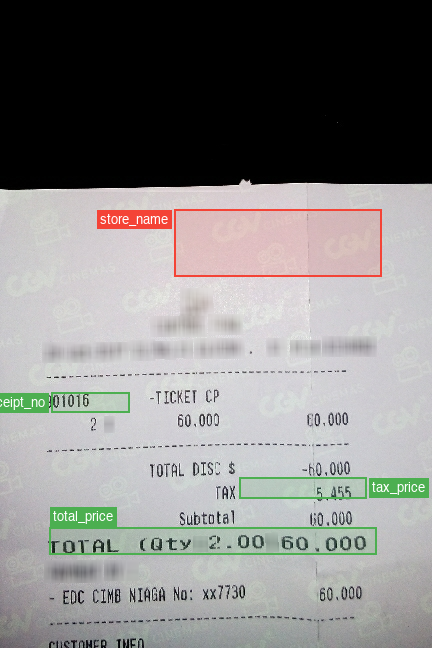}}\\[4pt]
\footnotesize
\begin{tabular}{@{}lll@{}}
\toprule
\textbf{Field} & \textbf{Extracted} & \textbf{Route} \\
\midrule
\rowcolor{green!8} total\_price & 60,000 & \textcolor{green!60!black}{$\bullet$~GREEN} \\
\rowcolor{green!8} tax\_price & 5,455 & \textcolor{green!60!black}{$\bullet$~GREEN} \\
\rowcolor{red!8} store\_name & Cinema & \textcolor{red}{$\bullet$~RED} \\
\rowcolor{green!8} receipt\_no & 901016 & \textcolor{green!60!black}{$\bullet$~GREEN} \\
\bottomrule
\end{tabular}\\[3pt]
\textbf{(a)} CORD v2 receipt
\end{minipage}
\hfill
\begin{minipage}[t]{0.48\textwidth}
\centering
\setlength{\fboxsep}{0pt}\setlength{\fboxrule}{0.4pt}%
\fbox{\includegraphics[width=\linewidth]{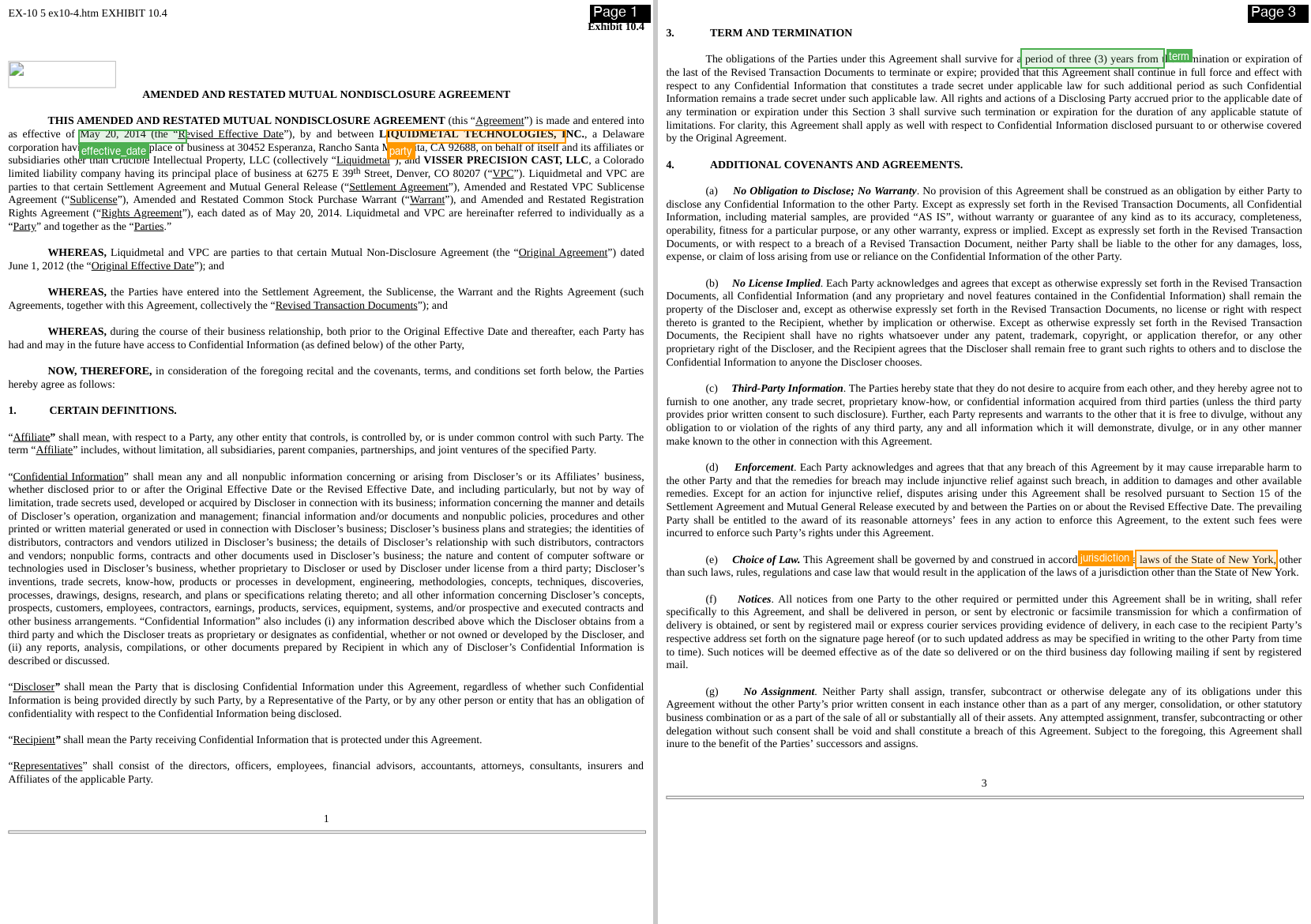}}\\[4pt]
\footnotesize
\begin{tabular}{@{}lll@{}}
\toprule
\textbf{Field} & \textbf{Extracted} & \textbf{Route} \\
\midrule
\rowcolor{green!8} effective\_date & 2014-05-20 & \textcolor{green!60!black}{$\bullet$~GREEN} \\
\rowcolor{orange!10} jurisdiction & New York & \textcolor{orange!80!black}{$\bullet$~AMBER} \\
\rowcolor{orange!10} party & Liquidmetal...; Visser... & \textcolor{orange!80!black}{$\bullet$~AMBER} \\
\rowcolor{green!8} term & 3 years & \textcolor{green!60!black}{$\bullet$~GREEN} \\
\bottomrule
\end{tabular}\\[3pt]
\textbf{(b)} Kleister NDA (pages 1 \& 3 of 4)
\end{minipage}
\caption{Example documents with per-field confidence routing.
Coloured bounding boxes on the images correspond to table rows.
\textcolor{green!60!black}{\textbf{GREEN}} fields are auto-accepted;
\textcolor{orange!80!black}{\textbf{AMBER}}/\textcolor{red}{\textbf{RED}}
fields are routed to human review at the \emph{field} level, not the
document level.}
\label{fig:examples}
\Description{Two example documents with extracted fields colour-coded by confidence: a CORD receipt and a Kleister NDA contract page, with bounding box overlays matching table row colours.}
\end{figure*}

\section{Experimental Setup}

\paragraph{Datasets.}
We evaluate on two public KIE benchmarks chosen to span different document
types and complexity levels.
\textbf{CORD~v2}~\cite{park2019cord} contains 100 test receipt images with 9
top-level scalar extraction fields (sub\_total, tax, total, store name, address,
cashier, receipt number, date, time); the full JSON schema includes additional
nested fields (e.g., menu item names and prices) that contribute to the
scoreable set when the model produces valid output for them.
\textbf{Kleister NDA}~\cite{stanislawek2021kleister} (dev-0 split) contains 83
multi-page (median~5 pages) Non-Disclosure Agreements filed with the SEC, with 4 fields
(effective date, jurisdiction, parties, term).
CORD represents a well-structured single-page domain; Kleister NDA represents
a harder multi-page legal domain relevant to financial operations
(Figure~\ref{fig:examples}).

\paragraph{Models.}
We evaluate three VLMs spanning different architectures, scales, and intended
use cases:
(1)~\textbf{Qwen3-VL-8B}~\cite{qwen3vl2025}, specifically the
\emph{Thinking} variant with chain-of-thought suppressed via chat-template
patching (Section~\ref{sec:thinking} handles residual thinking tokens),
an 8B parameter VLM supporting
native multi-image input;
(2)~\textbf{Granite-2.9B}~\cite{granite2025vision}, a 2.9B parameter
model built on LLaVA-Next, designed for enterprise document understanding;
(3)~\textbf{Gemma~4-5.1B}~\cite{gemma4report}, a 5.1B general-purpose multimodal
model from Google, not specifically designed for document extraction.
For models that accept only a single image (Granite, Gemma), we concatenate
multi-page documents vertically into one image.
All models run in bfloat16 with greedy decoding (temperature~0);
Section~\ref{sec:greedy} ablates this choice.

\paragraph{Baselines.}
We compare against two baselines: (1)~\emph{Verbalized confidence}, where the
model outputs a high/medium/low confidence label per field (mapped to
1.0/0.5/0.0); and (2)~\emph{Random}, with uniform random scores (AUROC~0.50).

\paragraph{Metrics.}
AUROC treats incorrect extractions as the positive class, with
$1 - \texttt{confidence}$ as the score.
AUARC measures selective prediction quality~\cite{geifman2017selective}
(sort by confidence, measure accuracy at each coverage level).
ECE measures probability calibration.
Extraction accuracy is normalised exact match (case-insensitive, whitespace-collapsed).

\section{Confidence Signal Quality}

\subsection{The Confidence Signal Predicts Errors}

Table~\ref{tab:rag} breaks down extraction accuracy by routing bin for Qwen3-VL
on CORD.
GREEN fields are correct 90.8\% of the time, more than double the accuracy of RED
fields (37.8\%).
The selective prediction curve achieves AUARC~=~0.942, meaning that reviewing
fields in order of decreasing confidence catches errors rapidly.
\emph{Note on alignment rate:} of the 900 top-level field slots (100
samples $\times$ 9 fields), 285 (32\%) receive aligned confidence scores.
The remaining 615 consist of fields where the model output ``NA'' (matching
unannotated ground truth) and fields where the JSON span scanner could not
locate the value.
Both cases receive RED by default and are excluded from calibration analysis.
On NDA, where all fields have ground-truth annotations, alignment is 100\%
on the 249 scalar-field extractions (3 scalar fields $\times$ 83 documents;
the list field \texttt{party} is excluded from scalar analysis).

\begin{table}[t]
\centering
\small
\begin{tabular}{lrr}
\toprule
\textbf{Routing Bin} & \textbf{$n$ fields} & \textbf{Accuracy} \\
\midrule
GREEN & 228 & \textbf{0.908} \\
AMBER &  12 & 0.333 \\
RED   &  45 & 0.378 \\
\midrule
Overall (scored) & 285 & 0.800 \\
\bottomrule
\end{tabular}
\caption{Extraction accuracy by confidence routing bin (Qwen3-VL-8B, CORD~v2,
greedy decoding). Only the 285 fields with successful logprob-to-span alignment
are included; a later experiment (Section~\ref{sec:cord-grpo}) evaluates the full JSON-parseable set.
AMBER accuracy is based on only 12 fields and should be interpreted cautiously.}
\label{tab:rag}
\end{table}

\begin{figure}[t]
\centering
\includegraphics[width=0.9\columnwidth]{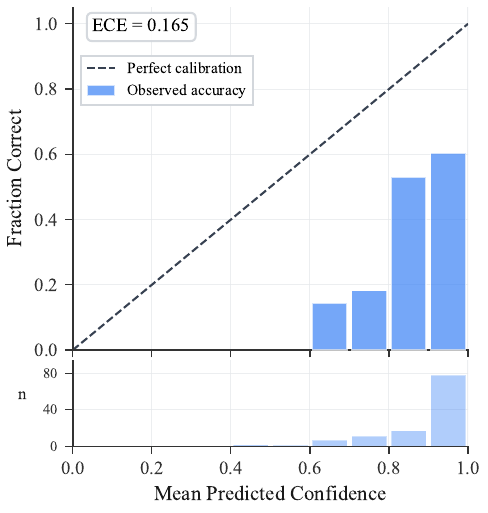}
\caption{Reliability diagram (Qwen3-VL-8B, CORD, greedy). ECE\,=\,0.165.
Bars show observed accuracy per confidence bin; the gap below the diagonal
indicates systematic overconfidence. Bottom: sample counts per bin.}
\label{fig:reliability}
\Description{Reliability diagram comparing predicted confidence to actual accuracy across ten bins, showing consistent overconfidence: observed accuracy falls well below the diagonal across all populated bins.}
\end{figure}

Figure~\ref{fig:reliability} shows the reliability diagram for Qwen3-VL on CORD.
The model is consistently overconfident: observed accuracy falls well below
predicted confidence across all populated bins, typical of instruction-tuned
models~\cite{nakkiran2025rl}.

\begin{figure}[t]
\centering
\includegraphics[width=0.9\columnwidth]{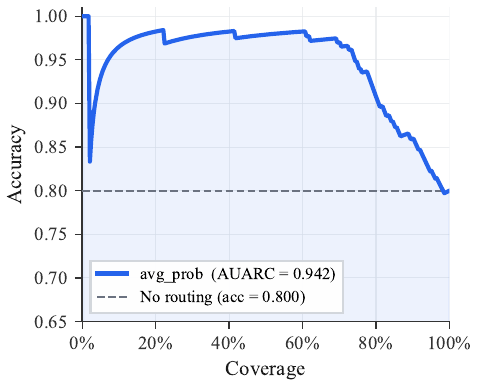}
\caption{Selective prediction curve (Qwen3-VL-8B, CORD). AUARC\,=\,0.942.
Reviewing in confidence order catches errors rapidly.}
\label{fig:selective}
\Description{Selective prediction curve showing accuracy rising steeply as coverage decreases, with AUARC of 0.942.}
\end{figure}

Figure~\ref{fig:selective} shows the selective prediction curve.
Sorting fields by descending confidence and accepting the most confident fraction
first yields high accuracy at low coverage, which is the practical operating mode for
production systems that process a fixed review budget per batch.

\subsection{The Signal Generalises Across Models and Domains}
\label{sec:cross}

A confidence signal that works only with one model on one dataset would be of
limited practical value.
Table~\ref{tab:multimodel} reports \texttt{min\_prob} AUROC across the full
3-model~$\times$~2-dataset grid.

\begin{table}[t]
\centering
\small
\begin{tabular}{lcccc}
\toprule
 & \multicolumn{2}{c}{\textbf{CORD (receipts)}} & \multicolumn{2}{c}{\textbf{NDA (legal)}} \\
\cmidrule(lr){2-3}\cmidrule(lr){4-5}
\textbf{Model} & \textbf{AUROC} & \textbf{Acc} & \textbf{AUROC} & \textbf{Acc} \\
\midrule
Qwen3-VL-8B    & \textbf{0.914}{\scriptsize$\pm$0.04} & 0.800 & \textbf{0.819}{\scriptsize$\pm$0.06} & 0.791 \\
Granite-2.9B   & 0.846{\scriptsize$\pm$0.03}          & 0.787 & 0.768{\scriptsize$\pm$0.05}          & 0.289 \\
Gemma~4-5.1B   & 0.809{\scriptsize$\pm$0.03}          & 0.335 & 0.745{\scriptsize$\pm$0.06}          & 0.188 \\
\midrule
Verbalized     & 0.563 & --- & 0.527 & --- \\
Random         & 0.500 & --- & 0.500 & --- \\
\bottomrule
\end{tabular}
\caption{\texttt{min\_prob} AUROC (with 95\% bootstrap CI half-width) across
three model families and two datasets.
All CIs exclude 0.5; the signal is statistically discriminative in every
condition.  Verbalized and Random baselines produce only a confidence
score, not extractions, so accuracy is not applicable (---).}
\label{tab:multimodel}
\end{table}

Three patterns emerge from the cross-model comparison.
First, AUROC remains well above random in all six cells, ranging from 0.745
(Gemma on NDA) to 0.914 (Qwen on CORD).
Second, the signal is discriminative even when extraction accuracy is poor:
Gemma~4 achieves only 33.5\% accuracy on CORD yet its \texttt{min\_prob}
still separates correct from incorrect predictions with AUROC~0.809.
On NDA, Granite and Gemma achieve very low extraction accuracy (28.9\%
and 18.8\%, as neither was trained on multi-page legal documents) yet the
confidence signal remains discriminative (AUROC 0.77 and 0.75).
Third, verbalized confidence (where the model is explicitly prompted to assess
its own certainty) achieves AUROC 0.53--0.56 on both datasets, essentially random.
On NDA, the model assigns ``high'' to 326 of 332 fields (including the list field
\texttt{party}) regardless of correctness.
This confirms that VLMs cannot reliably assess their own
extraction confidence when queried verbally, consistent with the findings
of~\citet{yang2024verbalized} in the text-only setting.

\subsection{Which Signal Works Best?}

Table~\ref{tab:signals} compares the four logprob-derived signals and their
combinations on CORD with the primary model (Qwen3-VL-8B).

\begin{table}[t]
\centering
\small
\begin{tabular}{lc}
\toprule
\textbf{Confidence Signal} & \textbf{AUROC} \\
\midrule
Composite (domain-weighted) & \textbf{0.937} \\
\texttt{min\_prob} & 0.914 \\
\texttt{avg\_entropy} & 0.905 \\
\texttt{geo\_prob} & 0.900 \\
\texttt{avg\_prob} & 0.898 \\
Rank-average ensemble & 0.899 \\
Verbalized (high/med/low) & 0.563 \\
\bottomrule
\end{tabular}
\caption{Error detection AUROC for individual signals and combinations
(Qwen3-VL-8B, CORD, greedy). The composite uses domain-calibrated weights;
the rank-average ensemble is parameter-free.}
\label{tab:signals}
\end{table}

\texttt{min\_prob}, the probability of the least-confident token in a field's
span, is the single best zero-shot signal (0.914).
This aligns with the structural finding of~\citet{snel2025firsttoken}: uncertainty
concentrates at the onset of an incorrect generation, and a single uncertain token
is more informative than the average across all tokens.
A domain-weighted composite that incorporates entropy, variance, and spread
achieves 0.937, a further 2.3-point improvement when domain weights are available.
A parameter-free rank-average ensemble (0.899) actually \emph{underperforms}
\texttt{min\_prob}, suggesting that the signals are highly correlated and
na\"ive averaging dilutes the strongest predictor.

\subsection{Cross-Dataset Results on Kleister NDA}

Table~\ref{tab:nda} reports the primary model's performance on the legal document
benchmark.
The NDA dataset is substantially harder than receipts: documents are multi-page,
field values are longer (party names, legal jurisdictions), and the relevant
information may appear on any page.

\begin{table}[t]
\centering
\small
\begin{tabular}{lcc}
\toprule
\textbf{Metric} & \textbf{CORD} & \textbf{NDA} \\
\midrule
Scored scalar fields & 285 & 249 \\
Overall accuracy & 0.800 & 0.791 \\
GREEN $n$ / accuracy & 228 / 0.908 & 168 / 0.905 \\
AMBER $n$ / accuracy & 12 / 0.333 & 35 / 0.629 \\
RED $n$ / accuracy & 45 / 0.378 & 46 / 0.500 \\
\midrule
AUROC (\texttt{min\_prob}) & \textbf{0.914} & 0.819 \\
AUROC (composite) & 0.937 & 0.819 \\
AUARC & 0.942 & 0.915 \\
ECE & 0.165 & 0.174 \\
\bottomrule
\end{tabular}
\caption{Qwen3-VL-8B performance on CORD vs.\ Kleister NDA.
GREEN accuracy exceeds 90\% on both; the fixed thresholds transfer without
recalibration.}
\label{tab:nda}
\end{table}

The central finding transfers: GREEN accuracy exceeds 90\% on both datasets, and
the same fixed thresholds (calibrated on a separate financial document corpus)
produce meaningful separation without domain-specific tuning.
The lower AUROC on NDA (0.819 vs.\ 0.914) reflects the different error structure:
NDA fields tend to be longer multi-token values where the model can be confidently
wrong on most tokens but uncertain on one or two, compressing the score range.
On NDA, the domain-weighted composite offers no improvement
(AUROC~0.819 for both), suggesting that the additional signals
(entropy, variance) add noise on long-field documents where
\texttt{min\_prob} already captures the dominant uncertainty mode.

\subsection{Greedy Decoding Is Essential}
\label{sec:greedy}

Table~\ref{tab:greedy} compares greedy and sampling decoding.
Every metric degrades under sampling, despite producing 2.2$\times$ more aligned
fields (the broader probability distribution keeps more tokens within the top-50
capture window).

\begin{table}[t]
\centering
\small
\begin{tabular}{lcc}
\toprule
\textbf{Metric} & \textbf{Greedy} & \textbf{Sampling ($T{=}1.0$)} \\
\midrule
Aligned fields & 285 & 636 \\
Overall accuracy & 0.800 & 0.665 \\
ECE ($\downarrow$) & \textbf{0.165} & 0.247 \\
AUARC ($\uparrow$) & \textbf{0.942} & 0.858 \\
AUROC (\texttt{min\_prob}) & \textbf{0.914} & 0.864 \\
GREEN accuracy & \textbf{0.908} & 0.876 \\
\bottomrule
\end{tabular}
\caption{Greedy vs.\ sampling on CORD (Qwen3-VL-8B). Sampling produces more
aligned fields but degrades every calibration metric. The temperature-scaled
distribution no longer reflects true uncertainty.}
\label{tab:greedy}
\end{table}

Under greedy decoding, a token with $p{=}0.95$ was selected \emph{because} it
had the highest probability---the logprob directly reflects model certainty.
Under sampling, a different token may be drawn; the captured logprob then
reflects what happened to be sampled, not what the model would deterministically
produce.
This validates our design: enforce greedy decoding for confidence estimation,
accepting a lower alignment rate for a faithful uncertainty signal.

\subsection{Few-Shot Recalibration}
\label{sec:recal}

The raw \texttt{min\_prob} signal has ECE~=~0.174 on held-out splits
(Table~\ref{tab:recal}; 0.165 on the full set per Table~\ref{tab:nda}),
useful for ranking but not for interpreting the score as a true probability.
We evaluate Platt scaling (logistic regression on the logit-transformed score) as
a post-hoc recalibrator, fitted on held-out subsets of varying size.
Table~\ref{tab:recal} shows the results.

\begin{table}[t]
\centering
\small
\begin{tabular}{rcccc}
\toprule
\textbf{$n$ labels} & \textbf{ECE (raw)} & \textbf{ECE (Platt)} & \textbf{GREEN acc (fixed)} & \textbf{GREEN acc (Platt)} \\
\midrule
 10 & 0.174 & 0.146 & 0.881 & 0.931 \\
 20 & 0.174 & 0.116 & 0.881 & 0.949 \\
 50 & 0.174 & 0.093 & 0.881 & 0.963 \\
100 & 0.174 & 0.084 & 0.882 & 0.968 \\
\bottomrule
\end{tabular}
\caption{Platt scaling on \texttt{min\_prob} (CORD, 500 random splits per size).
GREEN accuracy rises substantially; AUROC is unchanged (monotone transform).}
\label{tab:recal}
\end{table}

With just 50 labelled examples, Platt scaling reduces ECE by 47\% and raises
GREEN accuracy from 88.1\% to 96.3\% (averaged over random validation splits;
the full-set GREEN accuracy is 90.8\% per Table~\ref{tab:rag}).
AUROC is unaffected (0.914 before and after); calibration and discrimination are
orthogonal, and any monotone transform preserves ranking.
Temperature scaling, by contrast, \emph{worsens} ECE (to $\approx$0.33) because
the log-odds rescaling assumption does not hold for logprob signals.

\section{Confidence-Guided Self-Improvement}

Having established that the confidence signal reliably separates correct from
incorrect extractions, we now use it to drive model improvement.

\subsection{Single-Page Documents (CORD)}
\label{sec:cord-grpo}

Using the confidence-guided curriculum described in Section~\ref{sec:self},
we select 67 RED+AMBER field extractions from the CORD training split.
Table~\ref{tab:sft} compares SFT and GRPO on this curriculum.

\noindent\textbf{SFT} (supervised fine-tuning): cross-entropy loss on
(image, focused-prompt)~$\to$~correct-JSON pairs.
Training takes 17~minutes.
GREEN accuracy rises from 0.903 to 0.980; overall from 0.785 to 0.824.

\noindent\textbf{GRPO}: training takes 3.3~hours.
RED+AMBER collapses from 285 to 57 fields ($-$80\%; this 285 coincidentally
equals the aligned-field count in Table~\ref{tab:rag} but refers to
non-GREEN fields in the larger evaluation set); overall accuracy rises
to 0.875.

\begin{table}[t]
\centering
\small
\begin{tabular}{lccc}
\toprule
\textbf{Condition} & \textbf{GREEN $n$ / acc} & \textbf{A+R $n$} & \textbf{Overall} \\
\midrule
Base   & 402 / 0.903 & 285 & 0.785 \\
SFT    & 587 / \textbf{0.980} & 282 & 0.824 \\
GRPO   & 857 / 0.912 & \textbf{57} & \textbf{0.875} \\
\bottomrule
\end{tabular}
\caption{Self-improvement on CORD (67 training examples, rank-16 QLoRA).
The total field count exceeds the 285 scored fields in Table~\ref{tab:rag}
because the improved model now produces valid JSON for fields previously
left empty or unparseable (including nested sub-fields), expanding the
scoreable set.
Note: the Base A+R count of 285 is a coincidence---it refers to non-GREEN
fields from the larger evaluation set, not the 285 aligned fields in
Table~\ref{tab:rag}.
SFT maximises precision; GRPO maximises throughput.}
\label{tab:sft}
\end{table}

\begin{figure*}[t]
\centering
\begin{minipage}{0.48\textwidth}
\centering
\includegraphics[width=\textwidth]{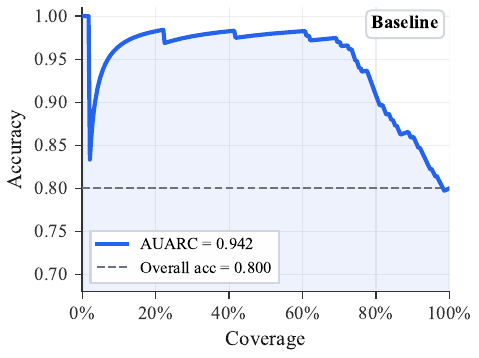}\\[2pt]
\small\textit{(a) Baseline (AUARC\,=\,0.942)}
\end{minipage}\hfill
\begin{minipage}{0.48\textwidth}
\centering
\includegraphics[width=\textwidth]{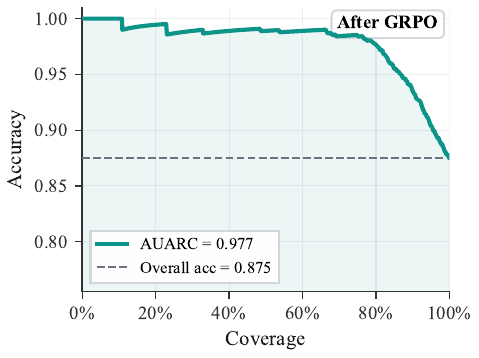}\\[2pt]
\small\textit{(b) After GRPO (AUARC\,=\,0.977)}
\end{minipage}
\caption{Selective prediction curves on CORD before and after GRPO
fine-tuning.  The post-GRPO curve is uniformly higher, reflecting both
improved accuracy (87.5\% vs.\ 78.5\% on the expanded scoreable set) and
better calibration (the model's confidence ranking aligns more closely
with actual correctness).}
\label{fig:cord_selective}
\Description{Two selective prediction curves for CORD: both show accuracy decreasing with increasing coverage, but the GRPO curve is uniformly higher than the baseline.}
\end{figure*}

SFT and GRPO favour different ends of the precision--throughput spectrum.
SFT achieves the highest GREEN precision (0.980), appropriate when false positives
in the automated path carry high cost (e.g., settlement of incorrect notional
amounts).
GRPO maximises throughput: 857 of 914 aligned fields route directly to GREEN,
reducing the human review burden by 80\%.
Both use the same 67 examples selected automatically by the confidence signal,
with no manual triage of training data.
Figure~\ref{fig:cord_selective} shows the selective prediction curves before
and after GRPO; the post-GRPO curve is uniformly higher, with AUARC rising
from 0.942 to 0.977.

\subsection{Multi-Page Documents (NDA)}
\label{sec:nda-grpo}
The confidence-guided curriculum extends to multi-page documents but
requires native multi-image training.
We apply GRPO on 259 RED+AMBER NDA field extractions from the train
split, using documents up to 8~pages with each page as a separate image.
Table~\ref{tab:nda_grpo} reports the results.
We use the same GRPO configuration as CORD (Section~\ref{sec:self}).
Because GRPO training uses only documents of up to 8~pages, we evaluate
both baseline and fine-tuned models on the same ${\le}8$-page subset for
a fair comparison.
The resulting baseline accuracy (70.7\%) is lower than the 79.1\%
reported in Table~\ref{tab:nda}, which uses all pages up to~10; fields
such as \texttt{jurisdiction} that appear on later pages are missed
under the 8-page cap.
Within this controlled setting, GRPO lifts accuracy by
4.4~percentage points and sharply improves calibration
(\texttt{min\_prob} AUROC~0.70$\to$0.85).

\begin{table}[t]
\centering
\small
\begin{tabular}{lcc}
\toprule
\textbf{Metric} & \textbf{Baseline} & \textbf{GRPO (8-page)} \\
\midrule
Overall accuracy & 0.707 & \textbf{0.751} \\
ECE ($\downarrow$) & 0.247 & \textbf{0.227} \\
AUARC ($\uparrow$) & 0.778{\scriptsize$\pm$.08} & \textbf{0.916}{\scriptsize$\pm$.04} \\
AUROC (\texttt{avg\_prob}, $\uparrow$) & 0.691{\scriptsize$\pm$.08} & \textbf{0.843}{\scriptsize$\pm$.06} \\
AUROC (\texttt{min\_prob}, $\uparrow$) & 0.697{\scriptsize$\pm$.08} & \textbf{0.845}{\scriptsize$\pm$.06} \\
\midrule
GREEN / AMBER / RED & 175 / 29 / 45 & 219 / 15 / 15 \\
\bottomrule
\end{tabular}
\caption{GRPO self-improvement on Kleister NDA (Qwen3-VL-8B,
259 training examples, rank-16 QLoRA, 3 epochs, native multi-image input).
All calibration metrics improve substantially; RED fields collapse from
45 to 15 ($-$67\%).
Both \texttt{avg\_prob} and \texttt{min\_prob} AUROC are reported; they
are nearly identical on NDA, unlike CORD where \texttt{min\_prob} leads
by 1.6 points (Table~\ref{tab:signals}).}
\label{tab:nda_grpo}
\end{table}

All four metrics improve.
The largest gain is in error detection: \texttt{min\_prob} AUROC rises
from 0.697 to 0.845, a 21\% relative improvement, meaning the model's
confidence signal becomes far more discriminative after GRPO training.
RED fields collapse from 45 to 15 ($-$67\%), and accuracy rises by 4.4
percentage points.

Table~\ref{tab:nda_perfield} breaks down the gains by field.
The improvement is concentrated in \texttt{jurisdiction} (+19.3pp, net
+16 fields), the hardest field---values typically appear on later pages
and were the primary source of RED classifications.
\texttt{term} improves modestly (+2.4pp), while \texttt{effective\_date}
regresses slightly ($-$8.4pp, net $-$7 fields), likely because the
curriculum is dominated by \texttt{jurisdiction} examples.
The calibration improvement, by contrast, is universal: the confidence
signal becomes more discriminative for all fields.

\begin{table}[t]
\centering
\small
\begin{tabular}{lccccc}
\toprule
\textbf{Field} & \textbf{Base} & \textbf{GRPO} & \textbf{$\Delta$} & \textbf{+} & \textbf{$-$} \\
\midrule
effective\_date & 0.819 & 0.735 & $-$8.4 &  2 &  9 \\
jurisdiction    & 0.723 & \textbf{0.916} & +19.3 & 19 &  3 \\
term            & 0.578 & 0.602 & +2.4 &  8 &  6 \\
\midrule
Overall         & 0.707 & \textbf{0.751} & +4.4 & 29 & 18 \\
\bottomrule
\end{tabular}
\caption{Per-field accuracy on NDA before and after GRPO (8-page evaluation).
$\Delta$: percentage-point change.
+/$-$: fields that switched from incorrect to correct or vice versa.
GRPO sharply improves the hardest field (\texttt{jurisdiction}) while
slightly degrading \texttt{effective\_date}; the net accuracy gain of 4.4pp
masks this asymmetry.}
\label{tab:nda_perfield}
\end{table}

\begin{figure*}[t]
\centering
\begin{minipage}[t]{0.48\textwidth}
\centering
\includegraphics[width=\textwidth]{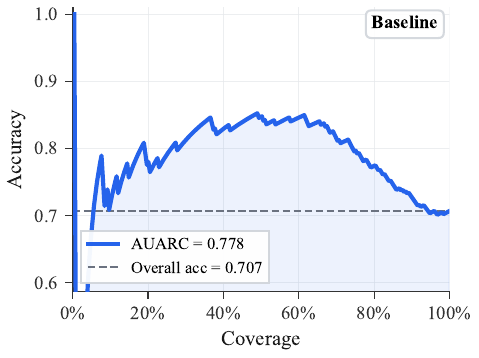}\\[2pt]
\small\textit{(a) Baseline (AUARC\,=\,0.778)}
\end{minipage}
\hfill
\begin{minipage}[t]{0.48\textwidth}
\centering
\includegraphics[width=\textwidth]{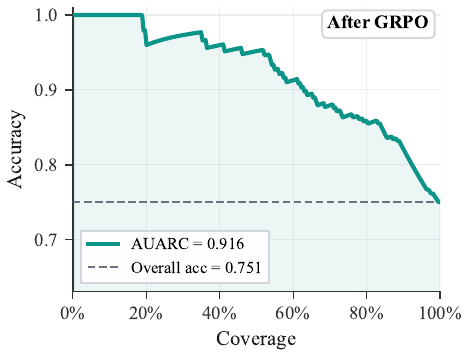}\\[2pt]
\small\textit{(b) After GRPO (AUARC\,=\,0.916)}
\end{minipage}
\caption{Selective prediction curves on Kleister NDA before and after
GRPO fine-tuning.
The baseline model's confidence signal is noisy and barely exceeds overall
accuracy at intermediate coverage; after GRPO, the curve is smooth and
maintains near-perfect accuracy up to 20\% coverage, enabling practical
selective-prediction workflows on multi-page legal documents.}
\label{fig:nda_selective}
\Description{Two selective prediction curves for NDA: the baseline curve is noisy; the GRPO curve is smooth with near-perfect accuracy at low coverage.}
\end{figure*}

Curriculum volume matters: an earlier run using only documents up to 5 pages
(163 examples) improved calibration (AUROC~0.829) but did not improve accuracy
(69.9\%); expanding to 8 pages provided the additional diversity needed for
accuracy gains.
Figure~\ref{fig:nda_selective} shows the selective prediction curves before
and after GRPO; the post-GRPO curve is markedly smoother and maintains
near-perfect accuracy up to 20\% coverage.

\paragraph{Engineering considerations.}
Applying GRPO to VLMs required addressing several challenges:
(1)~position-encoding mismatches when batching variable-length
image+text sequences;
(2)~suppressing the 1000+ token chain-of-thought reasoning that
thinking-mode models generate before the JSON answer, reducing
completions to 10--28 tokens;
(3)~setting $\beta{=}0$ (no KL penalty) to avoid reference-model
failures under 4-bit quantisation, relying on the LoRA constraint
for regularisation;
and (4)~native multi-image input during GRPO training, with each
page passed as a separate image.

\subsection{When Does Self-Improvement Work? The Role of Curriculum Design}
\label{sec:curriculum}

The self-improvement loop succeeds on both CORD and NDA, but the conditions
differ instructively.

The \emph{error rate within the selected curriculum} is one key factor.
On CORD, the RED+AMBER set (selected by \texttt{min\_prob}-based RAG thresholds)
has 63\% error rate: most selected fields are genuinely wrong, giving SFT/GRPO
clear targets. On NDA, the same selection yields only 32\% error: the model is
uncertain (low \texttt{min\_prob}) but usually \emph{correct}, because long
multi-token fields (party names, jurisdictions) naturally have at least one
token with low probability even in correct extractions.

Table~\ref{tab:curriculum} shows that the optimal signal for curriculum
selection depends on field length. On CORD (short fields, 3--5 tokens),
all signals select curricula with similar error rates because a single
uncertain token dominates the short span. On NDA (long fields, 10--40 tokens),
\texttt{min\_prob} is a weaker curriculum selector (32\% error) while
\texttt{avg\_prob} selects fields where the \emph{entire} value is uncertain
(52\% error at the bottom 100).

\begin{table}[t]
\centering
\small
\begin{tabular}{lcc}
\toprule
\textbf{Selection signal} & \textbf{CORD err\%} & \textbf{NDA err\%} \\
 & (bottom 50) & (bottom 100) \\
\midrule
\texttt{min\_prob} & 0.68 & 0.32 \\
\texttt{avg\_prob} & 0.62 & \textbf{0.52} \\
\texttt{geo\_prob} & 0.62 & \textbf{0.52} \\
Composite & 0.66 & 0.42 \\
\texttt{avg\_entropy} & 0.60 & 0.46 \\
\midrule
RAG RED+AMBER & 0.63 & 0.32 \\
\bottomrule
\end{tabular}
\caption{Curriculum error rate by selection signal. On CORD (short fields),
all signals are equivalent. On NDA (long fields), \texttt{avg\_prob} identifies
genuinely wrong fields while \texttt{min\_prob} selects uncertain-but-correct ones.
Higher curriculum error rate provides stronger learning signal, though
training format fidelity and data volume are equally important (see text).}
\label{tab:curriculum}
\end{table}

This finding resolves an apparent contradiction: \texttt{min\_prob} is the best
\emph{routing} signal (highest AUROC for ranking all fields) but a weaker
\emph{curriculum selector} for long-field documents. AUROC measures ranking
quality across the full score range; curriculum selection operates at the
low-confidence tail, where the two signals diverge. For short fields, the
distinction does not matter because \texttt{min} $\approx$ \texttt{avg} on 3--5 token spans.

However, curriculum error rate alone does not determine success.
Our NDA experiments (Table~\ref{tab:nda_grpo}) show that two additional
factors are critical: (1)~\emph{training format fidelity}---native multi-image
input during GRPO training, matching the multi-image evaluation format, rather
than concatenated single-image approximations; and (2)~\emph{curriculum
volume}---expanding from 5-page to 8-page documents increases the training set
from 163 to 259 examples, providing the diversity needed for accuracy gains.
With both conditions met, the loop produces substantial improvements on NDA
across all metrics (Table~\ref{tab:nda_grpo}).

The practical recommendation follows: use \texttt{min\_prob} for routing
(it achieves the best AUROC everywhere), but for curriculum selection on
long-field documents, select by \texttt{avg\_prob} $<$ threshold to ensure
the training set contains a majority of genuinely incorrect predictions.
(Our NDA GRPO experiment used RAG-threshold selection; the curriculum-analysis
above suggests \texttt{avg\_prob} selection could yield further gains.)
Ensure training format matches evaluation (native multi-image for multi-page
documents) and include sufficient curriculum volume.

\subsection{Toward Unsupervised Self-Improvement}
\label{sec:pseudo}

The self-improvement loop described above requires ground-truth labels for
the GRPO reward signal.
A fully unsupervised variant would replace these with pseudo-labels from a
frontier model.
To assess feasibility, we ran Claude Sonnet~4 as a judge on both
benchmarks, comparing its independent extractions against the same ground
truth used to compute GRPO rewards.

On CORD, Claude achieves 88.1\% extraction accuracy and its reward
signal agrees with the ground-truth reward 92.5\% of the time (67~fields,
1.5\% false-positive rate).
On NDA, exact-match accuracy is only 51.1\%---Claude extracts correct
content but in verbose form (e.g., ``State of Delaware'' vs.\ GT
``Delaware'')---yet with token-overlap matching (analogous to the
partial-F1 reward already in our GRPO setup), accuracy rises to 93.7\%
and reward agreement to 92.0\% (174~fields).
These results suggest that frontier-model pseudo-labels are a viable reward
signal for both document types, potentially enabling fully unsupervised
self-improvement on unlabelled document collections.

\section{Discussion}

\paragraph{Why does the minimum probability work so well for routing?}
Averaging masks the signal: in a 20-token span where 19 tokens have
$p{>}0.99$ and one has $p{=}0.3$, the arithmetic mean is still 0.96---safely
GREEN---while the minimum correctly flags the field as uncertain.
The effect is amplified for structured extraction, where most tokens are
\emph{format tokens} (quotes, commas, braces) that always receive high
probability; only the \emph{content tokens} carrying the actual value can
be uncertain, and the minimum isolates them.

\paragraph{Implications for financial document processing.}
The granularity of the signal changes the economics of review.
Document-level routing sends the entire document to a human reviewer whenever
\emph{any} field is uncertain; field-level routing sends only the uncertain
fields, which constitute 20\% of CORD outputs and 33\% of NDA outputs
at the fixed thresholds.
After GRPO on CORD, the absolute count of RED+AMBER fields drops from
285 to 57 ($-$80\%) while the scoreable set expands to 914 fields, leaving
only 6\% for human review.
For a pipeline processing 10k documents per day, this is the difference
between reviewing thousands of documents and reviewing hundreds of individual
field values---roughly an order of magnitude reduction in review labour.

\paragraph{Cross-model consistency.}
Across three architecturally distinct models, the signal works without modification.
We interpret this as suggestive evidence that the property is inherent to autoregressive
generation: when a model assigns low probability to a token it generates, that
probability reflects genuine uncertainty regardless of whether the model is a
document-specialist (Qwen), an enterprise tool (Granite), or a general-purpose
assistant (Gemma).
For financial institutions, the practical consequence is vendor independence.
Because the confidence layer operates on raw logprobs rather than model internals,
it transfers across model upgrades without revalidation.

\paragraph{Composite signals and domain adaptation.}
The domain-weighted composite (Table~\ref{tab:signals}) offers a 2--3 point
AUROC improvement over \texttt{min\_prob} alone, at the cost of requiring a
validation set to tune five weights.
For rapid deployment without domain labels, \texttt{min\_prob} is the
recommended default; for established pipelines with labelled validation
data, the composite is a free upgrade that requires no model changes.

\paragraph{Confidence-guided re-extraction.}
Not all RED fields are equally hopeless.
We investigate a two-pass protocol: for each RED scalar field, re-prompt the model
with a focused single-field instruction and check whether the original and
re-extracted values agree.
On CORD (188 RED fields), 39.6\% are consistent across passes; consistent RED
fields achieve 81.1\% accuracy versus 33.6\% for inconsistent ones
(a 2.4$\times$ separation).
This suggests a tiered workflow: accept GREEN; re-extract RED with a focused
prompt; promote consistent re-extractions to spot-check; route only
inconsistent fields to full human review.

\paragraph{Deployment considerations.}
The confidence computation adds negligible latency: the logprob hook runs during
generation (no extra passes), and the span alignment + JSON parsing takes
$<$10ms per document.
The method requires no GPU memory beyond what generation itself uses.
In a production deployment on financial documents, the pipeline runs as a
post-processing step after extraction, adding $<$1\% to end-to-end latency.
The routing thresholds can be adjusted per-client or per-document-type without
retraining: tightening the GREEN threshold increases precision at the cost of
sending more fields to review.

\paragraph{Error analysis: when does the signal fail?}
We manually inspect the 21 GREEN-but-wrong cases on CORD to understand failure
modes.
The majority (18/21) arise from annotation gaps in the CORD ground truth:
fields like \texttt{store\_name} and \texttt{receipt\_no} are unannotated for
many test samples, so a correctly extracted value scores as ``wrong'' against
the missing GT label.
Only 2--3 cases represent genuine overconfidence, where the model
confidently hallucinated an incorrect value (e.g., predicting ``11,070''
when the GT is ``19,557'') with near-perfect token probabilities.
In these cases every digit token receives high probability because the
model commits to an interpretation early and generates the remaining
digits consistently; the error is a global misread, not uncertainty at
any individual token.
The true GREEN overconfidence rate after adjusting for annotation gaps is
approximately 1\%, not 9\%.
When we exclude annotation-gap fields and recompute AUROC on the 243 fields with
confirmed ground truth, the corrected AUROC rises to 0.946 [0.901, 0.979],
indicating that the raw 0.914 is conservative.

On NDA, the GREEN-but-wrong cases have a different character: they tend to
involve jurisdiction or term fields where the model confidently outputs ``NA''
(correctly identifying that the field is not on the currently visible page)
while the GT expects a value from a later page.
This is a coverage problem rather than a calibration failure---the model is
genuinely certain the field is not visible, and it is right about that.

\paragraph{Post-training threshold recalibration.}
SFT and GRPO shift the score distribution, so the fixed thresholds no
longer sit at the optimal operating point.
Platt scaling on 50 labelled examples restores calibration and unlocks a
controllable precision--coverage trade-off without retraining
(Section~\ref{sec:recal}).
Recalibrate whenever the underlying
model changes---the cost is trivial compared to fine-tuning itself.

\paragraph{Limitations.}
Several limitations deserve mention.
First, while \emph{curriculum selection} is unsupervised, the GRPO
reward signal currently uses ground-truth labels; Section~\ref{sec:pseudo}
shows that frontier-model pseudo-labels achieve ${\sim}$92\% reward
agreement, but the effect on downstream GRPO training remains to be
validated.
Second, we evaluate a single GRPO round; iterating the loop may
compound gains but also risks overfitting to the confidence signal's
blind spots.
Third, all models are under 10B parameters; larger models may exhibit
different calibration properties.
Finally, the method assumes parseable structured output; when
generation fails entirely (invalid JSON), the field receives RED
by default without further discrimination.

\section{Conclusion}

We have shown that token log-probabilities, aligned to JSON field spans through
a lightweight post-hoc scanner, provide a strong and consistent confidence signal
for VLM-based document extraction.
Across two document domains and three model families, the minimum token
probability per field achieves AUROC 0.74--0.91, far exceeding verbalized
confidence (0.53--0.56) and enabling practical selective-prediction workflows
(AUARC~0.92--0.94).
The signal can be recalibrated with as few as 50 labelled examples to push GREEN
accuracy above 96\%, and the uncertain fields it identifies serve as a
self-improvement curriculum that works across both single-page and multi-page
documents: +9pp accuracy on CORD receipts and +4.4pp on Kleister NDA contracts,
with error detection on NDA improving substantially
(AUROC~0.70$\to$0.85).
Achieving these gains on multi-page documents required native multi-image GRPO
training and sufficient curriculum volume (259 examples from documents up to 8
pages), demonstrating that training format fidelity and data diversity are as
important as curriculum selection strategy.
Moreover, a frontier model used as a pseudo-label judge achieves
${\sim}$92\% reward agreement with ground truth on both benchmarks,
opening a path toward fully unsupervised self-improvement.

For financial institutions deploying VLMs in document processing, the method
provides immediate value as a routing layer (GREEN fields flow automatically;
AMBER/RED fields receive targeted human review at the field level), a low-cost
recalibration pathway (50 labelled examples push GREEN accuracy above 96\%),
and a framework for continuous model improvement through
uncertainty-guided fine-tuning.

A promising direction for future work is iterating the
loop---re-evaluating confidence after GRPO, selecting a new curriculum from
the updated uncertainty landscape, and retraining---subject to the caveats
discussed in Section~7.

\bibliographystyle{ACM-Reference-Format}
\bibliography{references}

\end{document}